\documentclass[journal]{IEEEtran} 

\usepackage{cite} 

\usepackage[pdftex]{graphicx}
\graphicspath{{img/pdf/}{img/jpeg/}{img/png/}} 
\DeclareGraphicsExtensions{.pdf,.jpeg,.jpg,.png} 

\usepackage{amsmath}

\usepackage{url}

\usepackage[latin1]{inputenc}

\usepackage{color}

\usepackage{array}

\newcolumntype{C}[1]{>{\centering\arraybackslash}p{#1}}

\usepackage{colortbl}

\usepackage{algorithmic}

\begin{document}

\title{Intentions of Vulnerable Road Users -- Detection and Forecasting by Means of Machine Learning}
\author{Michael Goldhammer, Sebastian Köhler, Stefan Zernetsch, Konrad Doll, Bernhard Sick, and Klaus Dietmayer 
\thanks{M. Goldhammer, S. Köhler, S. Zernetsch and K. Doll are with the Faculty of Engineering, University of Applied Sciences Aschaffenburg, Aschaffenburg, Germany (e-mail: \{michael.goldhammer, sebastian.koehler, stefan.zernetsch, konrad.doll\}@h-ab.de)}
\thanks{B. Sick is with the Intelligent Embedded Systems Lab, University of Kassel, Kassel, Germany (e-mail: bsick@uni-kassel.de)}
\thanks{K. Dietmayer is with the Institute of Measurement Control and Microtechnology, Ulm University, Ulm, Germany (e-mail: klaus.dietmayer@uni-ulm.de)}
}


\maketitle

\begin{abstract}
Avoiding collisions with vulnerable road users (VRUs) using sensor-based early recognition of critical situations is one of the manifold opportunities provided by the current development in the field of intelligent vehicles. As especially pedestrians and cyclists are very agile and have a variety of movement options, modeling their behavior in traffic scenes is a challenging task. In this article we propose movement models based on machine learning methods, in particular artificial neural networks, in order to classify the current motion state and to predict the future trajectory of VRUs.
Both model types are also combined to enable the application of specifically trained motion predictors based on a continuously updated pseudo probabilistic state classification. Furthermore, the architecture is used to evaluate motion-specific physical models for starting and stopping and video-based pedestrian motion classification.
A comprehensive dataset consisting of 1068 pedestrian and 494 cyclist scenes acquired at an urban intersection is used for optimization, training, and evaluation of the different models.
The results show substantial higher classification rates and the ability to earlier recognize motion state changes with the machine learning approaches compared to interacting multiple model (IMM) Kalman Filtering. The trajectory prediction quality is also improved for all kinds of test scenes, especially when starting and stopping motions are included. Here, 37\% and 41\% lower position errors were achieved on average, respectively.
\end{abstract}

\begin{IEEEkeywords}
road safety, vulnerable road users, movement modeling, intention recognition, motion classification, trajectory prediction, artificial neural networks
\end{IEEEkeywords}

\section{Introduction}

Due to the World Health Organization's status report on road safety, traffic accidents currently constitute the leading cause of death for young people aged 15\,--\,29 years. Moreover, they are also one of the most frequent causes among most other age classes \cite{WHO2015}. About half of those cases concern vulnerable road users (VRU), i.\,e., pedestrians, cyclists, and motorcyclists.
While the progress of active and passive safety functions in the last decades has steadily improved the protection of  car passengers, the protection of VRUs still remains a critical issue. Passive safety concepts such as helmets or energy absorbing vehicle design play an important role but often cannot avoid severe injuries, even at relatively low velocities within urban accident scenarios.
A unique opportunity to close this gap between vehicle and VRU safety is provided by the current progress in the fields of advanced driver assistant systems (ADAS) and autonomous driving: By predicting critical situations and thus being able to take active countermeasures in an early state, VRU accidents could be avoided or at least their consequences could be significantly reduced.
A fundamental task thereby is the creation of suitable VRU movement models. In particular for pedestrians and cyclists, a behavior prediction is challenging as they often are not using specific traffic lanes, can abruptly change their motion state (e.\,g., starting, stopping, bending in) and do not actively communicate their intention by turn indicators or brake lights. Considering future applications, they also hardly have an option to broadcast their motion state and intention via vechicle-to-everything (V2X) communication such as motor vehicles. 

Addressing those issues we propose and investigate novel VRU movement models based on machine learning techniques in combination with large-scale realistic training data from public traffic scenes. We use an offline learning approach meaning that we first train the models and then the model are evaluated without learning anymore. Our concept includes the detection of motion state changes in an early phase, the prediction of the future trajectory and the combination of both approaches. The models are designed with the goals of having low demands regarding prior knowledge and a high degree of independence from general conditions such as traffic and environment situations. So, only the VRU's trajectory observed in any global coordinate system by an appropriate sensor technique is sufficient as input data. Additional environment information, e.\,g., map data, is not required.
Based on the same data, also other state-of-the-art models using video-features and physical motion parameters are optimized and compared to the proposed approaches.

The remainder of the article is structured as follows: In Section \ref{RelWork}, an overview over the related work in the areas of pedestrian and cyclist motion modeling and prediction is given. Our approach, broken down into the individual main processing steps is presented afterwards in Section \ref{approach}. In Section \ref{testSiteDataSets}, the test site and the VRU datasets used for training and evaluation of the models are described. The performed experiments and results are discussed in Section \ref{expRes}, while a final conclusion is drawn and a brief outlook is given in Section \ref{conclusion}. \vspace{-3mm}

\section{Related Work} \label{RelWork}

Modeling pedestrian and cyclist movement has already been a task in different areas of research, e.\,g., biomechanics, physiotherapy and sports sciences. The main objective of those models is the analysis of basic movement parameters for understanding and improving motion sequences or their individual components for certain groups or activities, e.\,g., physically restricted persons or athletes in popular and competitive sports. In contrast, the main purpose of VRU movement modeling is forecasting the short-term behavior for a continuous analysis of the traffic surrounding. In this sector a certain number of studies concerning pedestrians, but hardly any works on bicyclists were published within the last years.

A frequently used technique for trajectory prediction in many applications is Kalman Filtering (KF), where physical state variables are assumed as constant whenever the state information cannot be updated by observation \cite{Bar-Shalom01}. Published approaches addressing pedestrians use constant velocity (CV), constant acceleration (CA), and constant turn rate (CT) models, or even combinations of these by interacting multiple model (IMM) filters \cite{Bertozzi04,Binelli05,Meuter08,Schneider13}. Kalman Filtering offers the advantage of low demands on prior knowledge and, therefore, it is suitable in many cases, e.\,g., small time horizons or walking/cycling with steady state velocity. However, as a matter of principle, it results in larger prediction errors whenever motion state changes appear and, thus, the basic assumption is violated, e.\,g., during starting and stopping.
Furthermore, the detection of motion transitions can also be performed with Kalman Filtering \cite{Bar-Shalom01}, which is an important aspect of VRU intention recognition, and can serve as basis to classify critical situations (e.\,g., a person suddenly steps on the road) or to choose suitable movement models. A common approach is the usage of the model probability of an IMM-KF with constant position (CP) and CV model to distinguish standing from walking motion \cite{Keller11}.
Another trajectory-based approach is presented by Wakim et al., where the four classes \textit{standing still}, \textit{walking}, \textit{jogging}, and \textit{running} are modeled by Hidden Markov Models (HMM) and updated based on the measured absolute and angular velocity \cite{Wakim04}.

Besides this exclusively trajectory-based approaches, additional sensor specific information, e.\,g., gathered from monocular cameras, stereo cameras, or LiDAR systems is used in other publications. Keller and Gavrila compare stopping motion detection by KF approaches to stereo-vision based methods using dense optical flow \cite{Keller14}. They also perform a trajectory prediction and show that the KF methods are outperformed in this case. Quintero et al. determine the joint positions of pedestrians on data acquired by a high-resolution stereo camera and a LiDAR system and characterize the movement via Gaussian Process Dynamical Models (GPDM) \cite{Quintero14,Quintero14_2}. They perform a state classification within the classes \textit{standing}, \textit{starting}, \textit{walking}, and \textit{stopping} as well as a trajectory prediction for a time horizon of one second.
The head pose and the associated direction of view are detected and tracked in camera images by Schulz and Stiefelhagen \cite{Schulz12}. They use this feature among others to recognize the intention of pedestrians to cross the road with different models, primary IMM filters, and Latent-dynamic Conditional Random Fields (LDCRF) \cite{Schulz15,Schulz15_2}. Kooij et al. use camera-based context information of the scene environment such as the distance of the pedestrian to the curb and his head pose (line of sight) in order to rate the criticality of situations. Using a novel Dynamic Bayesian Network, they perform an early recognition of motion state transitions \cite{Kooij14}.

In the present article, the work listed in this paragraph is extended, combined and extensively evaluated.
A trajectory prediction method based on two physical models of the pedestrian starting motion is published in \cite{Goldhammer13}, corresponding analyses for stopping motions in \cite{Goldhammer14}. A machine learning model using the time series of the pedestrian's ego velocity as input pattern for Multi Layer Perceptrons (MLP) is presented in  \cite{Goldhammer14_2}. This concept is extended with a polynomial representation by least-squares approximation of the input and output time series, leading to a reduction of the feature space dimensionality and an increase of generalisability \cite{Goldhammer15}. In \cite{Goldhammer16}, a modification of this model is used for the early recognition of the starting movement intention of pedestrians. A transfer of the approaches with physical and MLP models to cyclists is published in \cite{Zernetsch16}.
Video-based methods for motion classification and early recognition of state changes of pedestrians are presented in \cite{Koehler13} and \cite{Koehler15}. A Motion History Image (MHI) based histogram feature vector (MCHOG) approach is used as input for the classification of monocular images of a static camera \cite{Koehler13} and of stereo images from a moving vehicle \cite{Koehler15} by Support Vector Machines (SVM). \vspace{-2mm}

\section{Methodology} \label{approach}
\begin{figure}[!ht]
	\centering
	\vspace{-2mm}
	\includegraphics[width=1.0\columnwidth]{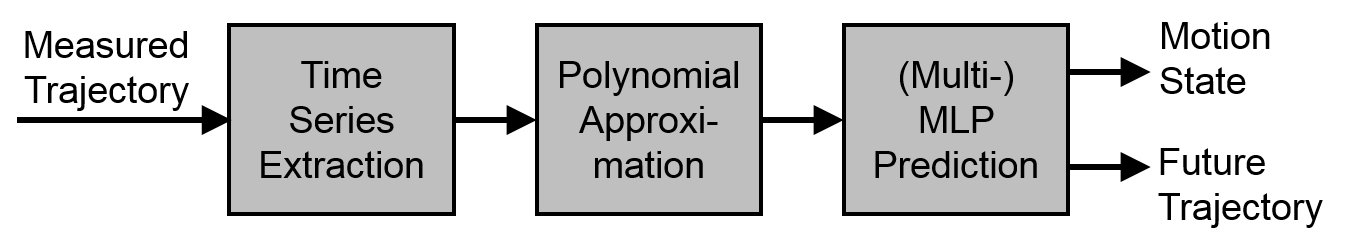}
	\caption{Overview of the basic processing steps of the method PolyMLP.}
	\label{fig:overview}
	\vspace{-2mm}
\end{figure}

For the recognition of the motion state we expand the method of polynomial approximation of time series and recognition of the starting movement (``PolyMLP'') presented in \cite{Goldhammer16} to a general classifier for the motion state. As input, we use the observed trajectory in a world coordinate system. As target output, a pseudo class posterior probability of the four states \textit{Waiting}, \textit{Starting}, \textit{Moving}, and \textit{Stopping} is provided. The method is compared to CP/CV-IMM-KF classification as baseline and to the MCHOG/SVM approach as solely video based classifier.
The PolyMLP method is also used for trajectory prediction. An overview of the input/output behavior is given in Fig.~\ref{fig:overview}. The resulting position accuracy is compared to CV-KF as baseline method and to physical models for starting and stopping optimized on the same training data.

In a final step, we combine the methods of motion state recognition and trajectory prediction to use motion state dependent optimized movement prediction models (e.g. specific PolyMLP only trained with starting movements) and evaluate their benefits compared to the single-stage (``monolithic'') PolyMLP.

Within this section, the successive steps of extracting characteristic time series (Sec.~\ref{tsExtraction}), their representation by polynomial least-squares approximation (Sec.~\ref{polynomials}), the prediction by MLP (Sec.~\ref{MlpPred}) as well as the target outputs of the motion state (Sec.~\ref{RecMovStat}) and the future trajectory (Sec.~\ref{PredTraj}) are described. Afterwards, in Sec.~\ref{SSTP} the motion state specific trajectory prediction model using multiple specifically trained MLP is presented.

\subsection{Time Series Extraction} \label{tsExtraction}
The PolyMLP method solely uses the information within the observed VRU trajectory in a static world coordinate system for the prediction of the motion state and/or the future trajectory. 
Therefore, it is very flexible with regard to the applied sensor system, but also allows for an easy extention of the input space with additional information.

For the acquisition of the trajectories evaluated within this study, an infrastructure based wide-angle stereo camera system and, alternatively, an array of multilayer laser scanners is used (see Sec.~\ref{testSiteDataSets}). As anchor point for tracking in 3D world space serves the VRUs' centers of gravity (COG) determined from the laser scanner generated point cloud, or the center of the head detected from video data (see \cite{Goldhammer14_2}), respectively. As changes of the motion state, especially the begin of the pedestrian starting motion, are initialized by a slight upper body bending leading to a shift of the COG into the direction of movement \cite{Shiozawa04}, the head movement can serve as an early indicator for the intention.

To extract a time series representing the VRU motion independent from the absolute position and direction, several approaches were implemented and compared, e.\,g., the  two-dimensional trajectory or velocity in the pedestrians ego coordinate system or the combination of the absolute and angular velocity. The finally applied configuration uses the velocity $v_{lon}$, $v_{lat}$ as numerical differentiation of the position $X$ at the discrete time steps $k$, $k-1$, $k-2$, etc. in the VRU's ego coordinate system $x_{lon}$, $x_{lat}$
(see Fig.~\ref{fig:poly} a, b).
Both dimensions of the time series are additionally processed with a first order exponential smoothing filter
\begin{equation}\label{smoothing}
S_k=\alpha y_{k}+(1-\alpha)S_{k-1},
\end{equation}
where $S_k$ is the current element of the time series, $y_{k}$ the current observation, $S_{k-1}$ the previous filtered element and $\alpha$ the smoothing factor. The values of $\alpha$ for both dimensions are part of the parameter set varied to optimize the overall prediction quality.

\subsection{Representation with Approximating Polynomials} \label{polynomials}
\begin{figure}[!ht]
	\vspace{-3mm}
	\centering
	\includegraphics[width=1.0\columnwidth]{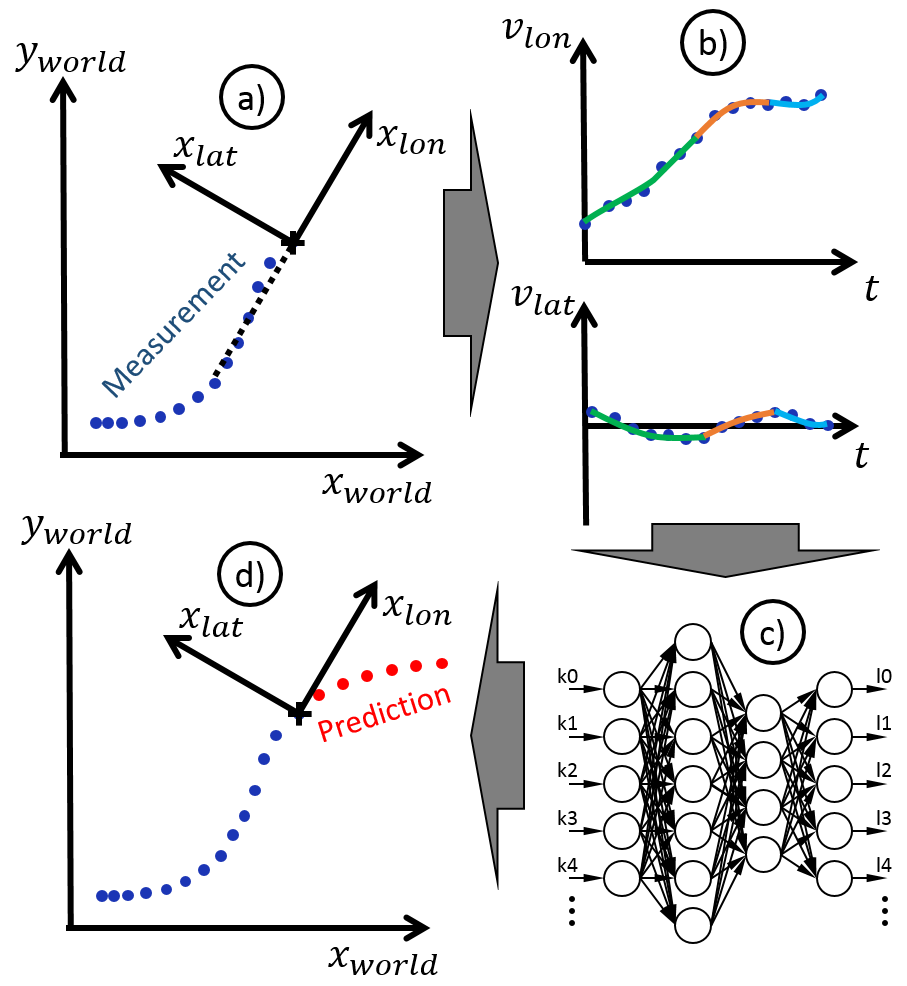}
	\vspace{-3mm}
	\caption{Overview of the proposed path prediction method.}
	\label{fig:poly}
	\vspace{-5mm}
\end{figure}

As shown in \cite{Goldhammer14_2}, the time series extracted within a certain time window can directly serve as input pattern of a machine learning predictor. However, we proceed to a further level of abstraction by using the coefficients of an orthogonal expansion of an approximating polynomial  as presented in \cite{Goldhammer16}.
The advantages of this step are a high grade of independence of the input data cycle time and a significant reduction of the dimensionality of the feature space. Dependent on the polynomial degree and the length of the time window, the approximation also reduces the influence of measurement noise. As the coefficients of the orthogonal expansion are optimal estimates for the average values of the time series' derivations, they can be interpreted as mean position, velocity, acceleration, etc. during the considered time window. Using the polynomials and update algorithms for sliding window processing of time series presented by Fuchs et al. \cite{Fuchs10}, the approximation can be performed very efficiently.
In order to allow the machine learning prediction algorithm for weighting different time periods separately as it is possible with direct input of the time series elements, we make use of multiple polynomials per dimension fitted in sequential temporal sub windows. Their number, temporal position, length and polynomial degree are model parameters and varied within the optimization. A schematic example with three sub windows is shown in Fig.~\ref{fig:poly} b (green, orange and blue polynomials).

\subsection{Multilayer Perceptron for Prediction} \label{MlpPred}
We apply an MLP to predict the motion state and/or the future trajectory of the VRU based on the extracted patterns (see Fig.~\ref{fig:poly} c). The MLP provides high flexibility and efficiency for the given application as it allows predicting multiple output dimensions at the same time. The applied artificial neural network consists of neurons with sigmoid activation functions, whereas other configurations (identity, Gaussian) were evaluated, too.
After normalizing the input patterns using a $z$-transformation, a training is performed with the Resilient Backpropagation (RPROP) algorithm \cite{Riedmiller93}. The number and sizes of the hidden layers are variable and also part of the optimization process.

\subsection{Recognition of Motion State} \label{RecMovStat}
	
\begin{figure}[!ht]
	\centering
	\includegraphics[width=.85\columnwidth]{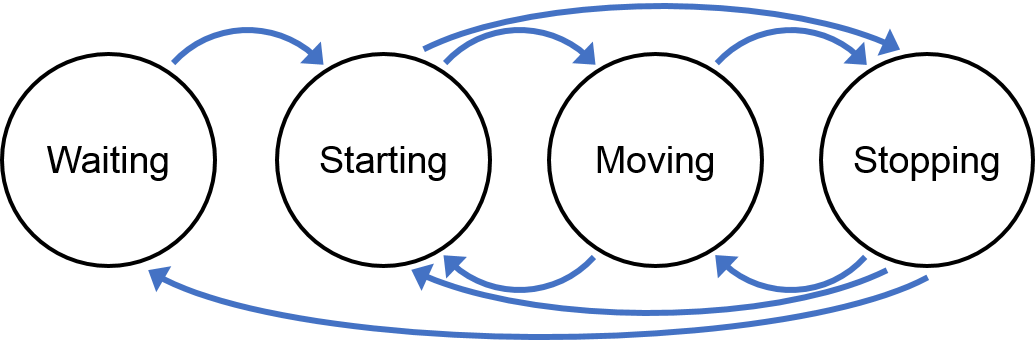}
	\caption{Modeled motion states and physically possible state transitions.}
	\label{fig:states}
\end{figure}

For the recognition of the motion state, a model of the four states \textit{Waiting}, \textit{Starting}, \textit{Moving}, and \textit{Stopping} is used. This basically allows the 8 (out of a possible basic quantity of 12) state transitions depicted in Fig.~\ref{fig:states}, as the transitions between \textit{Waiting} and \textit{Moving} are always separated by one of the transition states \textit{Starting} or \textit{Stopping}.
For the online classification, an MLP output layer with 4 neurons, one representing each state, is configured. The trajectories used for training are labeled with a distinct class label for every element, generating output training patterns with a one at the element representing the current ground truth class and zeros at all others.
No further measures are taken to define physically meaningful transitions or transition probabilities explicitly, as the neural network should be given the possibility to learn these aspects from scratch without further conditions.
With this approach, the trained classifier provides a pseudo-probabilistic rating for each state. As the four output neurons have no direct interconnection the cumulative probability is not explicitly specified to 100\%, so an additional scaling step has to be applied if this output information is needed by the subsequent application.






\subsection{Prediction of Trajectory} \label{PredTraj}
To predict the VRU trajectory, we use a representation according to the one of the input patterns.
To train the predictor, again characteristic time series are extracted from the ground truth trajectory within a certain time window (cf. Section~\ref{tsExtraction}) and the time series are approximated using orthogonal basis polynomials to get a low-dimensional representation independent of the cycle time (cf. Section~\ref{polynomials}). The choice of the MLP output pattern is completely independent of the one of the input pattern, what allows different time series and polynomial degrees.
Each polynomial coefficient is estimated by a separate MLP output neuron, where also a $z$-transformation is used for normalization.
During online application, the process is performed vice versa: The output pattern predicted by the MLP is transformed back and the time series is evaluated at the requested discrete points of time. It is then transformed to an estimation of the future trajectory in the original global coordinate system using the current VRU position and movement direction (see Fig.~\ref{fig:poly} d).

\subsection{State Specific Trajectory Prediction} \label{SSTP}
\begin{figure}[!ht]
	\centering
	\includegraphics[width=1.0\columnwidth]{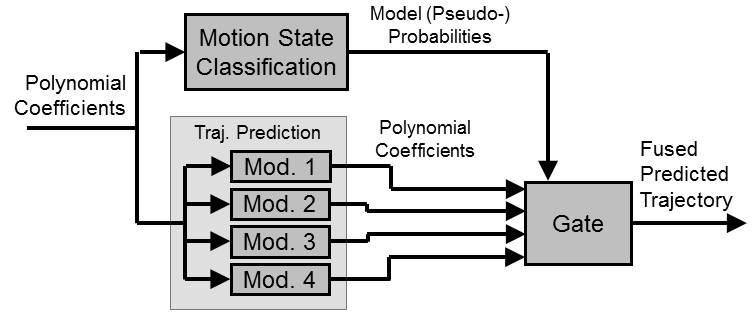}
	\caption{Motion state specific trajectory prediction.}
	\label{fig:twoStage}
	\vspace{-4mm}
\end{figure}

After training, the weights of the MLP implicitly contain the knowledge of the trained movements (starting, stopping, etc.) within a single model.
Besides this monolithic approach, a second approach combining state classification with several specifically trained models for trajectory prediction is implemented and evaluated. The architecture of the prediction system using the PolyMLP method is shown in Fig.~\ref{fig:twoStage}.

The measured trajectory is fed into a classifier module that predicts the current motion state as described in Sec.~\ref{RecMovStat} (Recognition of Movement State). In parallel, the trajectory is fed into several (here: four) PolyMLP modules for trajectory prediction. Each predictor is specifically trained with scenes containing movements according to one output of the classifier model, i.e. motion state. As the prediction modules can use the same configuration for preprocessing the trajectory and polynomial approximation, only the output of the specifically trained MLPs has to be processed separately. The predicted polynomial coefficients are fed into a gating module and added as weighted sum based on the class posterior probabilities. Finally, the resulting fused coefficient set is used to generate the final trajectory prediction.
This method of explicit separation of classification and trajectory prediction has the advantage that the classification as well as the prediction modules can be replaced by other approaches. For our evaluation we are thus able to use the MCHOG/SVM approach within the classification module or physical starting and stopping models in the prediction module as comparison techniques.

		
		
		

\section{Test Sites and Data Sets} \label{testSiteDataSets}

\begin{figure}[!ht]
		\centering
		\includegraphics[width=1.0\columnwidth]{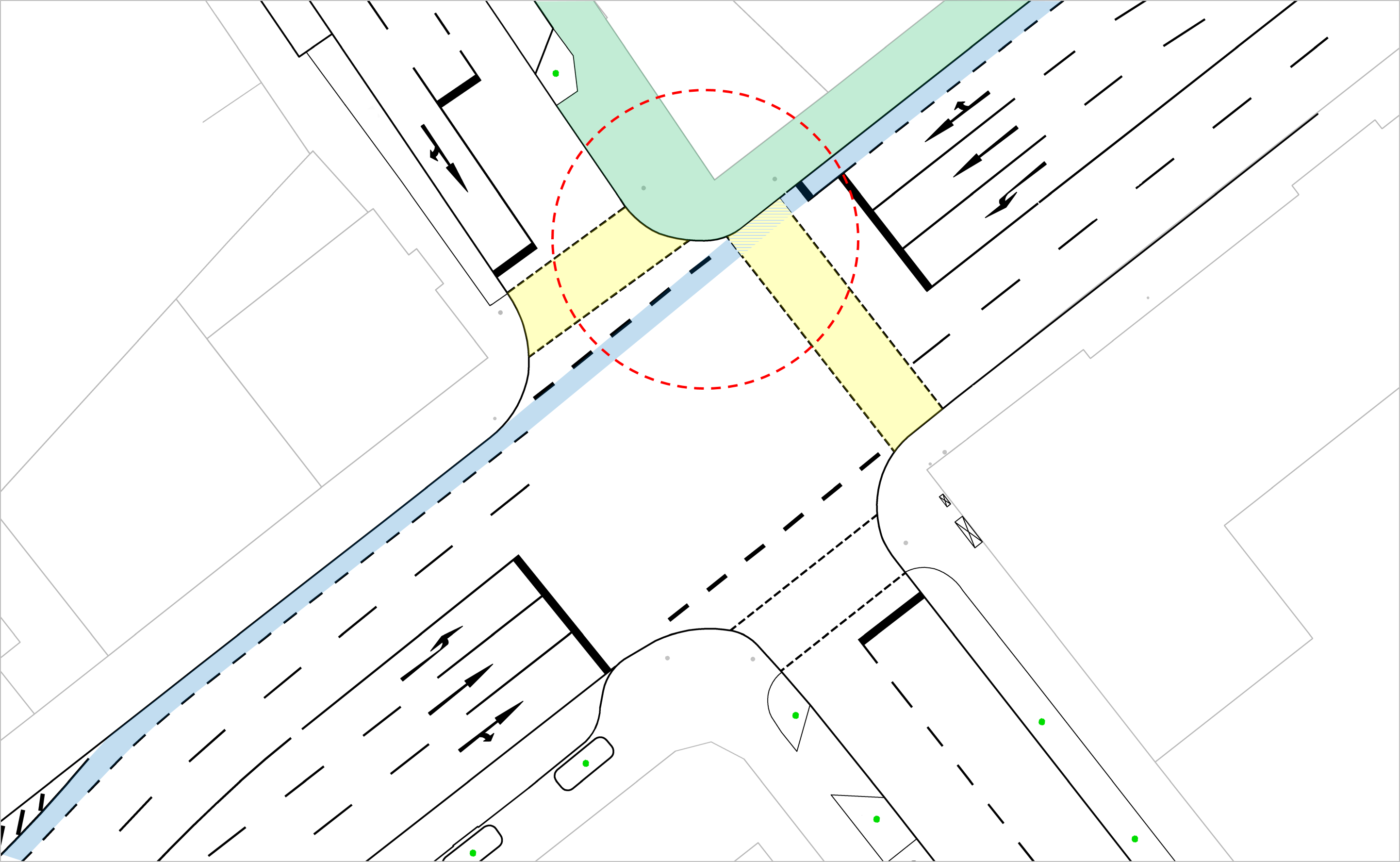}
		\caption{Map of the public intersection used to acquire VRU behavior data. The area highlighted in green represents the observed sidewalk, the yellow areas the crosswalks. The bicycle lane is highlighted in blue. The area of overlapping fields of view of the high resolution cameras allowing high-precision 3D position measurements is marked by the red circle.}
		\label{fig:intersection_map}
	\end{figure}

For training, optimization, and evaluation of the models, appropriate databases containing realistic VRU data play an essential role.
	We use an urban traffic intersection in order to gain measurements of realistic, unaffected pedestrian and cyclist behavior under real conditions. The observed area includes two sidewalks, two crosswalks and a bicycle lane which is separated from the motor vehicle lanes by road markings (see Fig.~\ref{fig:intersection_map}).
	The intersection is equipped with eight low-resolution (640\,x\,480\,px) and two high-resolution (1920\,x\,1080\,px) grayscale cameras, as well as 14 eight-layer laser scanners. All sensors are mounted on infrastructure elements, e.\,g., street lamps and traffic light posts, in heights between 4 and 11 meters in order to reduce the risk of occlusions. The sensor system is described in detail in \cite{Goldhammer12}.
	The high-resolution cameras are pointing towards a part of the intersection where two crosswalks meet and therefore a high quantity of pedestrians is expected (Fig.~\ref{fig:intersection_map}, dashed red circle). They form a wide angle stereo system allowing precise 3D measurements with an accuracy better than 3\,cm for corresponding image points. The laser scanners cover the central intersection and the area of three approaching roads up to 100\,m. They provide object point clouds and are used to track cyclists in a larger area beyond the stereo range of the camera system. The low resolution cameras are currently only used for manual scene observation and the labeling of movement phases.
	
\begin{figure}[!ht]
			\centering
			\includegraphics[width=1.0\columnwidth]{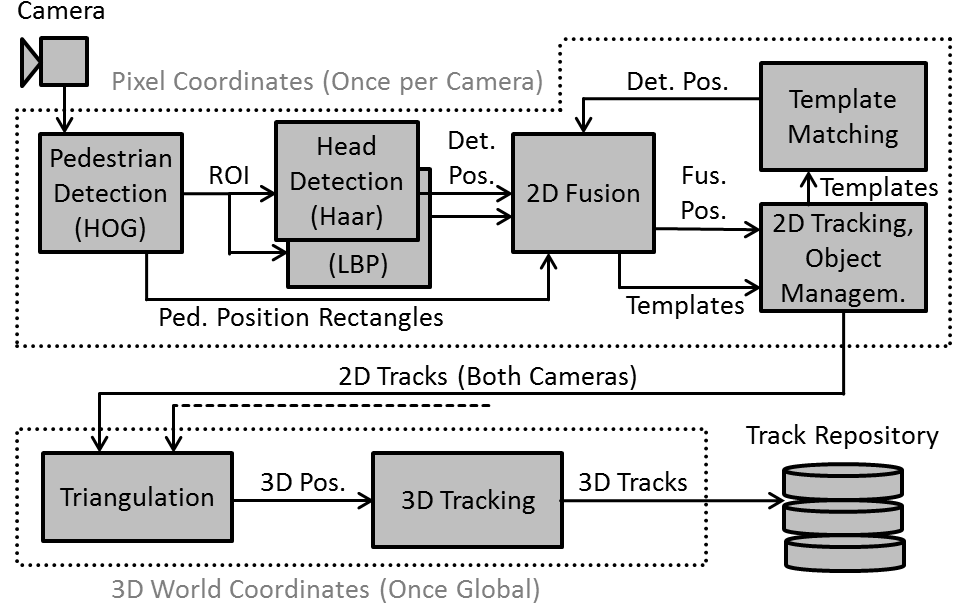}
			\caption{Framework for extraction of trajectories.}
			\label{fig_headTrackFW}
	\end{figure}

	Fig. \ref{fig_headTrackFW} shows the framework for the extraction of pedestrian and cyclist head trajectories from stereo video data. At the first stage, both synchronous camera frames are individually scanned by a sliding-window histogram of oriented gradients (HOG) pedestrian detector \cite{Dalal.2005}. The upper half of the detection rectangles serve as region of interest (ROI) for the head detectors , which are based on Haar \cite{Viola.2001} and local binary pattern (LBP) features \cite{Liao.2007}. In the next step, the most likely head position in pixel coordinates is determined fusing HOG, Haar, and LBP detections. If the head position is already tracked, template matching is used as a fourth measurement value that is fed into the fusion module. The tracked head positions from both cameras are merged to 3D world coordinate positions via triangulation. In a final step, valid positions are connected to 3D trajectories and stored.

	\subsection{Full Pedestrian Dataset}
	The \textit{Full Pedestrian Dataset} contains 1068 scenes of pedestrians with 4 to 10 seconds length recorded by the high resolution stereo system with a sample rate of 50\,Hz. The scenes are categorized with one of the following scene labels (the number in brackets indicates the number of scenes): \begin{itemize}
		\item \textit{Moving scenes} (288) contains pedestrians crossing the observed area without stopping or significant deceleration, including scenes of walking along the sidewalk, crossing one of the roads, walking straight ahead or making a turn.
		\item \textit{Waiting scenes} (259) contains pedestrians standing at the sidewalk, mostly waiting for a green pedestrian light signal (head velocity lower than 0.3\,m/s).
		\item \textit{Starting scenes} (336) contains pedestrians accelerating from standing position. These scenes also contain up to 3 seconds before and after the actual acceleration phase if available within the data.
		\item \textit{Stopping scenes} (185) contains scenes where moving pedestrians decelerate to standstill. These scenes also contain up to 3 seconds before and after the actual deceleration phase.
	\end{itemize}
	The \textit{Starting} scenes include time labels of the start, and if existent within the scene, also of the end of the acceleration phase. The start is determined as the time where the ground truth head velocity exceeds 0.2\,m/s, its end is labeled at the first local maximum after the velocity exceeds 80\% of their steady state value. The labeling of \textit{Stopping} is done vice versa. Table \ref{tab:FullPedDB} gives an overview over the training and test scenes and extracted patterns (combination of predictor in- and output, generated for each time cycle with complete in- and output time window around).
	\begin{table}[!htbp]
		\caption{Training- and test data in the Full Pedestrian Dataset. The table shows the number of scenes used and of the according extracted patterns in brackets.}
		\vspace{-5mm}
		\begin{center}
			\begin{tabular}{|p{0.8cm}|C{1.4cm}|C{1.4cm}|C{1.4cm}|C{1.4cm}|}
				\hline & \textbf{Waiting} & \textbf{Starting} &\textbf{Moving} & \textbf{Stopping}\\
				\hline \textbf{Train} & 177 (54.6\,k) & 239 (66.5\,k) & 201 (45.2\,k) & 130 (42.6\,k) \\
				\hline \textbf{Test} & 82 (25.3\,k) & 97 (25.1\,k) & 87 (20.1\,k) & 55 (17.7\,k) \\ 
				\hline \textbf{Total} & 259 (79.8\,k) & 336 (91.6\,k) & 288 (65.3\,k) & 185 (60.3\,k) \\ 
				\hline 
			\end{tabular} 
		\end{center}
		\label{tab:FullPedDB}
	\end{table}
	
	\subsection{Detailed Pedestrian Dataset}
	The \textit{Detailed Pedestrian Dataset} is an extended subset of 136 \textit{Starting}, 107 \textit{Stopping}, and 69 \textit{Moving} scenes from the full set with finer differentiated ground truth motion states. It contains manually determined labels by video data observation, e.\,g., the state, the timestamps of heel-off, the first and second heel-down, the entering of the roadway and partial occlusions.
	
	\subsection{Cyclist Dataset}
	The \textit{Cyclist Dataset} contains 494 trajectories of bicyclists including 86 \textit{Moving}, 133 \textit{Waiting}, 197 \textit{Starting} and 78 \textit{Stopping} scenes. As the cyclists' phases of acceleration and deceleration typically take more time and distance as those of pedestrians, the laser scanner system was used to capture also tracks leaving the more limited common field of view of the high resolution stereo cameras. In return, a lower spatial and temporal resolution was accepted for those tracks.
	
	As well the pedestrian as the cyclist dataset with labeled head trajectories is available to the scientific community~\cite{VRUDataset}. 
	

\section{Experiments and Results} \label{expRes}


\subsection{Training and Optimization} \label{sec:train}
The training and optimization of the pedestrian models is performed using the training split (747 scenes) of the \textit{Full Pedestrian Database}. The training scenes are split again into 70\% for the MLP training with RPROP and 30\% as validation set for the optimization of the meta parameters (pre-filtering parameters, sliding window positions and sizes, polynomial representation). The target function of the MLP training is a minimization of the mean squared error (MSE) of the $z$-normalized output, while the meta parameters are optimized regarding a maximum accuracy (state classification) and a minimum average specific average Euclidean error (ASAEE, see Sec.~\ref{GenTrajPred}). For the selection of the input time series representation (Sec.~\ref{tsExtraction}) a five-fold cross validation within the training set is applied.

The results show that the time series of the 2D velocity in the pedestrians ego system perform best as basis for the input patterns of both state classification and trajectory prediction, while only decent differences to other tested representations (see Sec.~\ref{tsExtraction}) were observed (0\%\,--\,3\% for state clas., 3\%\,--\,9\% for traj. pred.). The representation with polynomials shows slight advantages compared to the direct input of each time series (0\%\,--\,5\%). The final outcome of the optimization of the sliding window and polynomial configuration are two consecutive time windows of 800\,ms and 200\,ms length with polynomial degrees of 3.
For the output pattern of the trajectory prediction, the time series of the 2D position in the ego system with five consecutive sliding windows (500\,ms each, polynomial degree 2) is used, leading to an overall prediction time horizon of 2.5\,s.

\subsection{Motion State Classification}
In the first experiment, the quality of the proposed method for motion state classification is evaluated. For comparison, an IMM-KF using CP/CV models and the directly video-based method MCHOG/SVM \cite{Koehler15} are applied to the same scenes. As the evaluation requires the additional labels of the \textit{Detailed Pedestrian Dataset} (i.e. the time of heel-off to distinguish between standing and starting phase), the following tests are performed using these patterns.
In the experiment, the overall quality of the different motion state classifiers is evaluated by means of the accuracy (ACC) and the F$_1$ score.

As the quality measures and both comparison methods only support binary classification,  the ground truth label \textit{Waiting} is assigned to timestamps before and the label \textit{Moving} to those during and after the manually determined heel-off within the 40 \textit{Starting} scenes.
Two PolyMLP predictors are trained: The first uses a four-classes output and a binarization after the prediction step by thresholding the sum probability $P_\text{Sum}$ of the three estimates for \textit{Starting}, \textit{Moving} and \textit{Stopping}: $P_\text{Sum}=P(\text{Start})+P(\text{Mov})+P(\text{Stop})$.
The second is directly trained with this two-class split and uses a single output neuron to predict $P_\text{Sum}$ while the class \textit{Waiting} is defined by the complement probability $1-P_\text{Sum}$.
The 35 \textit{Stopping} scenes are also tested with a four-class model, where a threshold on $P_\text{Sum}=P(\text{Stop})+P(\text{Wait})$ is used for binarization. The two-class model is again trained directly with this split and predicts the output with a single neuron. 
As the classifier quality measures are depending on the binarization threshold, the optimal value is chosen by maximization of the accuracy on the training data (an optimization of the F$_1$-score leads to the same operating point). 
The resulting quality measures of both test scene types using optimal bias values are shown in Table~\ref{tab:ClassRes}.
\begin{table}[tb]
	\caption{Accuracy (ACC) and F$_1$ score of the classifiers for the 40 starting and 35 stopping test scenes of the detailed pedestrian database.} 
	\vspace{-5mm}
	\begin{center}
		\begin{tabular}{|p{2.0cm}|C{.9cm}|C{.9cm}|C{.9cm}|C{.9cm}|}
			\hline & \multicolumn{2}{c|}{\textbf{Start}} & \multicolumn{2}{c|}{\textbf{Stop}}\\
			& \textbf{ACC} & \textbf{F$_1$} & \textbf{ACC} & \textbf{F$_1$}\\
			\hline \textbf{IMM-KF} & 0.9803 & 0.9565 & 0.9248 & 0.9548 \\ 
			\hline \textbf{PolyMLP (4 Cl.)} & 0.9819 & 0.9601 & 0.9381 & 0.9644\\
			\hline \textbf{PolyMLP (2 Cl.)} & 0.9824 & 0.9612 & 0.9363 & 0.9636\\
			\hline \textbf{MCHOG} & 0.9844 & 0.9659 & 0.9141 & 0.9503\\
			\hline 
		\end{tabular} 
	\end{center}
	\label{tab:ClassRes}
\end{table}
For the regarded \textit{Starting} scenes, the quality measures show only minor differences with slight advantages for MCHOG and slight disadvantages for the IMM-KF compared to PolyMLP. Clearer differences can be observed for the \textit{Stopping} scenes: Here, PolyMLP performs best, while IMM-KF outperforms MCHOG. The optimization thereby results in a relatively high threshold for the CV model probability $P_\text{CV}$ of the IMM-KF of 98.9\%, which means that stopping can be recognized in an early state. 

As the PolyMLP method performs a multi-class prediction of all four motion states, it is additionally evaluated using the 321 test tracks of the \textit{Full Pedestrian Dataset}. The result shows an accuracy of 88.6\%, which is lower than the one of the \textit{Detailed Pedestrian Dataset} as the classifier has now to distinguish between \textit{Starting}, \textit{Walking}, and \textit{Stopping}, too. More detailed results are set out in the confusion matrix in Table~\ref{tab:confusMat} which shows an obvious substantial asymmetry  of the classification rates of the four types of movement.  
\begin{table}[tb]
	\caption{Confusion matrix for the classification of the motion state. The rows contain the specific ground truth label. The columns contain the percentage of detections with regard to the total number of patterns of the ground truth label (sum of every row equals 100\%).} 
	\vspace{-5mm}
	\begin{center}
		\begin{tabular}{|p{1.0cm}|C{1.0cm}|C{1.0cm}|C{1.0cm}|C{1.0cm}|}
			\hline & \textbf{Waiting} & \textbf{Starting} & \textbf{Walking} & \textbf{Stopping} \\
			\hline \textbf{Waiting} & \cellcolor{yellow}98.6\% & 0.7\% & 0.0\% & 0.7\%\\ 
			\hline \textbf{Starting} & 11.8\% & \cellcolor{yellow}77.1\% & 8.8\% & 2.3\%\\
			\hline \textbf{Walking} & 2.0\% & 4.8\% & \cellcolor{yellow}88.1\% & 5.0\%\\ 
			\hline \textbf{Stopping} & 2.1\% & 2.2\% & 34.8\% & \cellcolor{yellow}60.9\%\\ 
			\hline 
		\end{tabular} 
	\end{center}
	\label{tab:confusMat}
\end{table}
The state \textit{Waiting} can be separated best from the three others, leading to a correct classification rate of 98,6\%. Most challenging is the transition from \textit{Walking} to \textit{Stopping}: Here, especially the begin of the stopping phase is difficult to distinguish from velocity variations during walking, crossing one of the roads, or bending in. 

\subsection{Early Recognition of State Transitions}

\begin{figure}[!ht]
	\centering
	\includegraphics[width=.9\columnwidth]{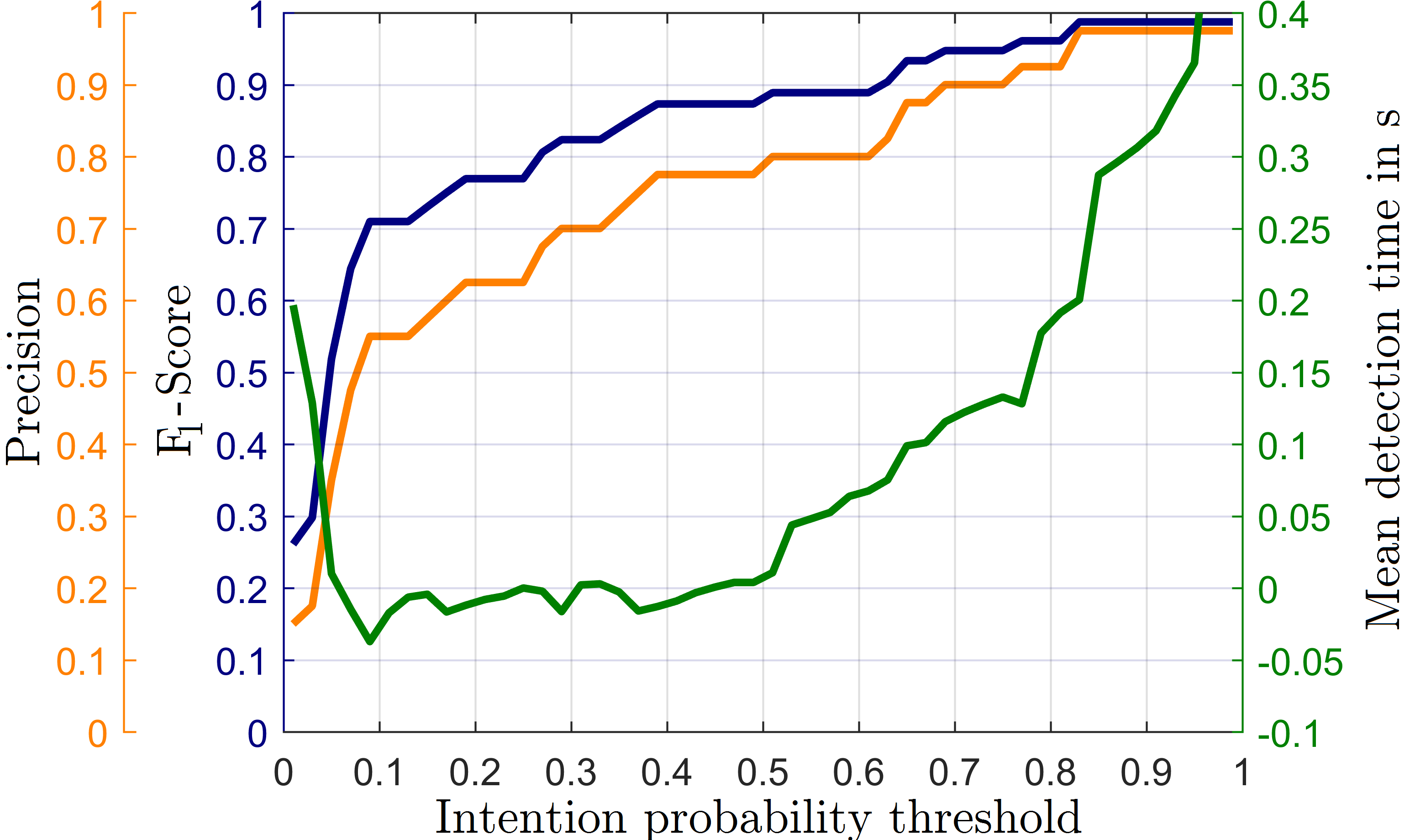}
	\includegraphics[width=.9\columnwidth]{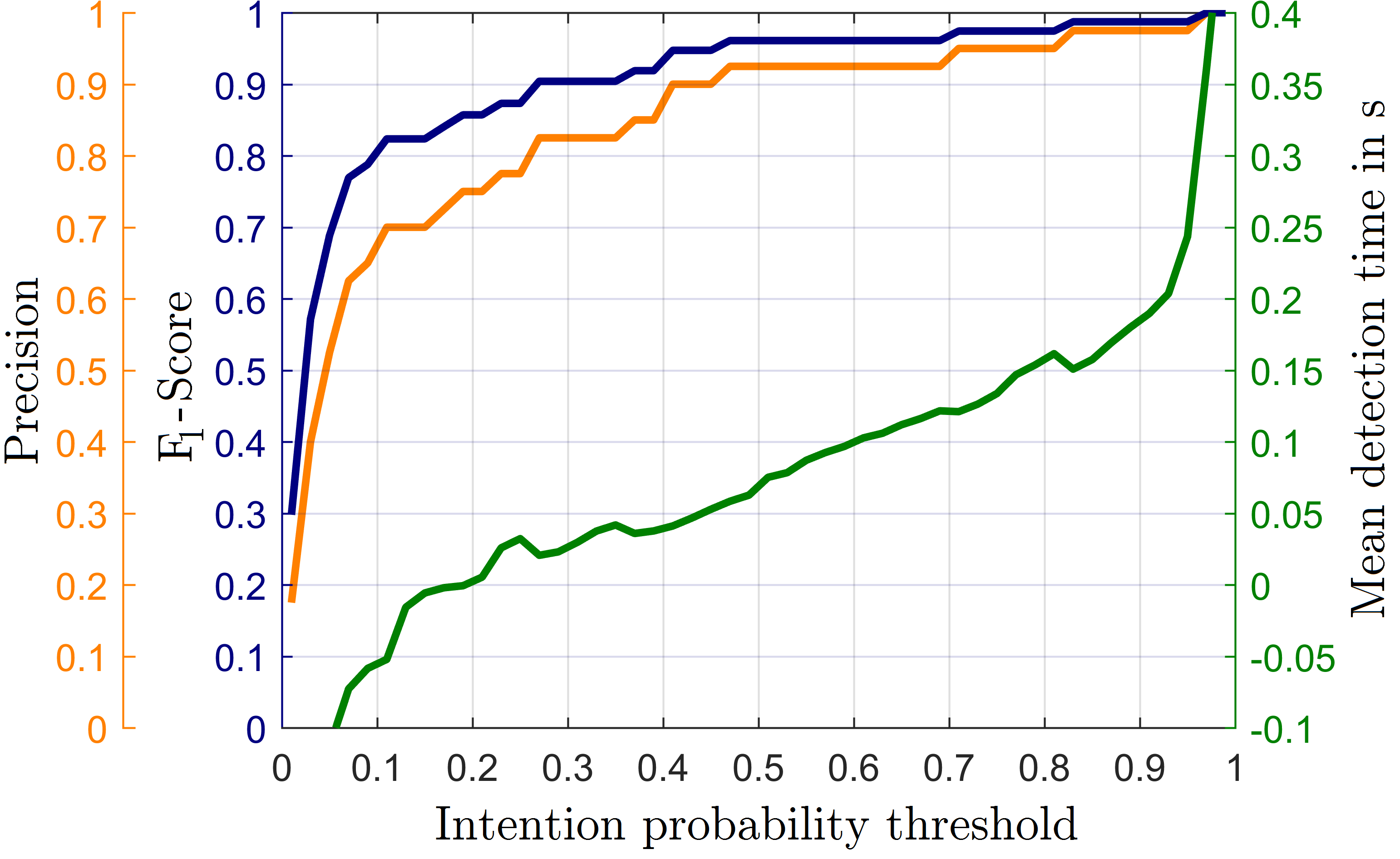}
	\caption{Evaluation of the recognition of starting movements using 40 scenes of the \textit{Detailed Pedestrian Dataset} for PolyMLP (upper plot) and MCHOG (lower plot). The plots show the precision (orange), the F$_1$-score (blue) and the mean time needed for detection of the starting motion relative to the manually labeled time of the heel-off (green). The chosen threshold on the pseudo probabilistic classifier output is drawn on the horizontal axis.}
	\label{fig:start_eval}
\end{figure}

For an evaluation of the prediction model ability to early recognize state transitions, the temporal development of the classification quality depending on the binarization threshold is analyzed. Here, the trade-off between the classifier's sensitivity leading to short reaction times on movement transitions on the one hand and low false alarm rates on the other hand should be examined.
For this purpose, the \textit{Detailed Pedestrian Dataset} is evaluated scene-wise on false positives under variation of the chosen threshold value. A scene is already considered as false positive if the classifier produces one false positive output at any single time step. Under that condition, the quality measures of the precision
and the F$_1$ score can be evaluated depending on the threshold. The additional determination of the accuracy does not make sense here as the number of true negatives is zero for most of the possible values (The state transition is detected earlier or later in any case.).
Furthermore, the mean time for the correct classification of the movement transition relative to the according manually labeled point in time is determined. For this evaluation, only scenes without false positives for the regarded threshold value are considered.
Fig.~\ref{fig:start_eval} shows the four quality measures evaluated on the \textit{Starting} scenes dependent on the chosen threshold for the PolyMLP and the MCHOG/SVM classifier.
The results show that an F$_1$ score of 95\% and a precision of 90\% are reached 130\,ms after the labeled heel-off. The MCHOG classifier is about 3 frames faster at the same operating point, and thus able to detect the starting motion about 70\,ms earlier. In contrast, the IMM-EKF needs 160\,ms (plot not shown).

\begin{figure}[!ht]
	\centering
	\includegraphics[width=.9\columnwidth]{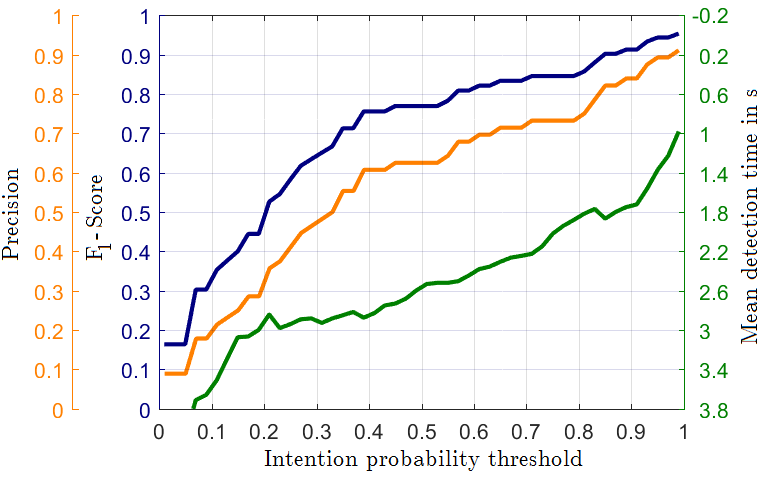}
	\includegraphics[width=.9\columnwidth]{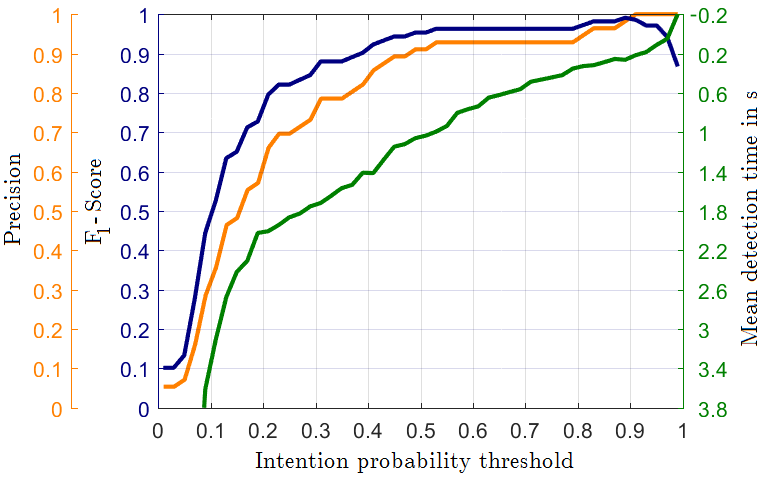}
	\caption{Evaluation of the recognition of stopping movements using 35 test scenes of the \textit{Detailed Pedestrian Dataset} for PolyMLP (upper plot) and MCHOG (lower plot). The axes and plot configuration correspond to Fig.~\ref{fig:start_eval}.}
	\label{fig:stop_eval}
\end{figure}

Fig.~\ref{fig:stop_eval} shows the plots for the same kind of evaluation regarding the recognition of stopping motion. Here, the detection time is measured in relation to the heel-down of the last step.
In this case, the PolyMLP method performs best, reaching an F$_1$ score of 95\% and a precision of 90\% already 1.4\,s before the heel-down (on average). Both other classifiers, MCHOG/SVM and IMM-KF, reach this quality level only 400\,ms later, which is a large time span in the field of active accident prevention.

Actually, we want to reach 100\% F$_1$ score and precision, but there is a tradeoff between F$_1$/precision and mean detection time and sometimes we are not able to get at 100\% F$_1$/precision. Therefore, we choose the 95\% F$_1$ and 90\% precision levels for comparison of the PolyMLP and the MCHOG/SVM classifier on starting motions (Fig.~\ref{fig:start_eval}) and stopping motions (Fig.~\ref{fig:stop_eval}). 
When analyzing accidents it was shown that an initiation of an emergency break 160 ms earlier at a Time-to-Collision of 660 ms and a vehicle speed of 50 km/h reduces the probability of an injury resulting in a hospital stay from 50\% to 35\%~\cite{Keller11}. This underlines the need for detecting basic movements as early as possible and that some milliseconds can make the difference. 
Fig.~\ref{fig:start_eval} shows that for an F$_1$ score/prediction  of 100\% some starting motions can only be detected 0.4 sec or later after heel off. This indicates that there is still research to be done.



\subsection{Generally Applicable Methods of Trajectory Prediction} \label{GenTrajPred}

In this section, the generally applicable methods of trajectory prediction, the monolithic and the two-stage specific PolyMLP, are evaluated and compared to CV Kalman filtering as baseline. As performance measure, the average specific average Euclidean error
\begin{equation}
\text{ASAEE}=\frac{1}{N}\sum\limits_{i=1}^{N}\frac{\text{AEE}(t_{pred}(i))}{t_{pred}(i)}
\end{equation}
with the AEE being the average Euclidean position error for a specific time span $t_{pred}(i)$, the index $i$ and the total number $N$ of discrete time spans predicted into the future (see \cite{Goldhammer16}). In our case, $N$\,=\,125 steps with $t_{pred}$\,=\,0.02\,..\,2.50\,s are used. The ASAEE is evaluated separately for the methods and the four scene types within the \textit{Full Pedestrian Dataset}, leading to the results shown in Table~\ref{tab:ResUniversal}.
 
The results of the Kalman Filter already show the varying degrees of difficulty of the single scene types: As expected, \textit{Waiting} generates the lowest errors (7.8\,cm/s). The ASAEE rises with increasing presence of velocity changes resulting in the further order \textit{Walking}, \textit{Stopping}, and \textit{Starting}. \textit{Starting} includes the most abrupt velocity changes in combination with the absence of a defined direction regarding the trajectory at the beginning of the movement.
\begin{table}[!tbp]
	\caption{ASAEE (in cm/s) of universally applicable prediction methods for the four scene classes in comparison. The table shows the results for CV Kalman Filtering (CV-KF), the single stage (monolothic) MLP model (PolyMLP), the two-stage MLP model (PolyMLP (2\,St.)) and the combination of the ground truth class labels with the specifically trained PolyMLP models (GT+PolyMLP). } 
	\vspace{-5mm}
	\begin{center}
		\begin{tabular}{|p{1.95cm}|C{.9cm}|C{.9cm}|C{.9cm}|C{.9cm}|C{.55cm}|}
			\hline & \textbf{Waiting} & \textbf{Starting} & \textbf{Walking} & \textbf{Stopping} & \textbf{Mean}\\
			\hline \textbf{CV-KF} & 7.8 & 44.2 & 27.6 & 33.5 & \textbf{28.3} \\ 
			\hline \textbf{PolyMLP} & 6.9 & 33.6 & 25.5 & 22.7 & \textbf{22.2}\\
			\hline \textbf{PolyMLP (2\,St.)} & 4.7 & 34.4 & 23.4 & 25.2 & \textbf{21.9}\\
			\hline \textbf{GT+PolyMLP} & 3.9 & 32.3 & 22.8 & 22.5 & \textbf{20.4}\\ 
			\hline 
		\end{tabular} 
	\end{center}
	\label{tab:ResUniversal}
\end{table}
For \textit{Waiting} and \textit{Walking}, the monolithic PolyMLP shows moderate improvements of 12\% and 8\% to the Kalman Filter. The improvement for scenes including movement transitions is substantially larger: \textit{Starting} scenes improve by 24\%, \textit{Stopping} scenes even by 32\%. Considering only the predictions directly within the labeled actual transition phase, an improvement of 42\% (\textit{Starting}) resp. 43\% (\textit{Stopping}) can be observed.

In comparison to the monolithic PolyMLP, the two-stage approach shows a slight further improvement of 1\,--\,2\,cm/s on \textit{Waiting} and \textit{Walking} scenes while the error for \textit{Starting} and \textit{Stopping} increases by approximately the same values. This means that the additional classification error outweighs the improvement of the specific trajectory prediction models for the transition scenes. Overall, the two-stage model leads to slightly lower errors, whereas the difference is only 1\% considering the average of the four scene class results and 3\% considering the ASAEE evaluated over all test patterns.

In order to examine the maximum potential of the two-stage approach, the classification stage is replaced by the ground truth class labels. Thus, the trajectory prediction stage is always able to choose the optimal model. The experimental results show a potential limit at a 7\% lower average scene class error. Considering the separate scene types, especially the \textit{Starting} and \textit{Stopping} prediction of the monolithic model already performs very close to the optimum.


\subsection{Two-stage Trajectory Prediction}

In this test series, the quality of the motion specific two-stage trajectory prediction methods for \textit{Starting} and \textit{Stopping} scenes are evaluated and compared to the generally applicable methods tested in the previous section using the \textit{Detailed Pedestrian Database}. We evaluate the predictions only during the labeled movement transition, as the dataset contains longer phases of waiting resp. walking before the starting resp. the stopping movement. The third scene class \textit{Moving} is included in order to evaluate the sensitivity for false alarm stopping classifications.

The results presented in Table~\ref{tab:ResSpecific} show the superiority of the machine learning models compared to the prediction of the (generally optimized) CV Kalman Filter. The application of the monolithic PolyMLP model leads to a reduction of the prediction error by 36.5\% for \textit{Starting} and 40.7\% for \textit{Stopping} scenes. As shown in the last row of the table by combining the ground truth classification with the specifically trained PolyMLP prediction, the potential for improvement by two-stage modeling is only 4.9\% for \textit{Starting}, but 11.0\% for \textit{Stopping} scenes. As classification stage we evaluated the four-class PolyMLP classifier already tested on the \textit{Full Pedestrian Dataset} (PolyMLP) and the image-based two-class MCHOG/SVM classifiers (MCHOG). For trajectory prediction, specifically trained PolyMLP models and physically based models for starting and stopping (PhysMod) are evaluated.
In the case of \textit{Starting}, the specific PolyMLP models outperform the physical model by 9\% while they perform almost equal on \textit{Stopping} motions. A distinct disadvantage of the physical \textit{Stopping} model thereby is its sensitivity on false alarms of the classifier stage, leading to a particularly high prediction error on \textit{Moving} scenes. 
Considering the machine learning models, it is remarkable that no two-stage approach is able to outperform the single-stage PolyMLP. While with \textit{Starting} scenes no substantial differences between the single-stage PolyMLP and the two-stage models with PolyMLP- and MCHOG classification are measured, the monolithic approach slightly outperforms both at \textit{Stopping} by 6.4\% resp. 3.7\%. \vspace{-3mm}
\begin{table}[!htbp]
	\caption{ASAEE (in cm/s) of the different optimized trajectory prediction methods for the three scene classes in comparison.} 
	\vspace{-5mm}
	\begin{center}
		\begin{tabular}{|p{2.6cm}|C{1.0cm}|C{1.0cm}|C{1.0cm}|}
			\hline & \textbf{Starting} & \textbf{Stopping} & \textbf{Moving} \\
			\hline \textbf{KF} & 49.94 & 36.04 & 23.35 \\ 
			\hline \textbf{PolyMLP} & 31.71 & 21.37 & 22.41 \\
			\hline \textbf{PolyMLP+PhysMod} & 35.23 & 21.75 & 41.97 \\
			\hline \textbf{PolyMLP+PolyMLP} & 31.54 & 22.84 & 23.10 \\ 
			\hline \textbf{MCHOG+PolyMLP} & 31.85 & 22.20 & 22.72 \\ 
			\hline \textbf{GT+PolyMLP} & 30.17 & 19.01 & 21.43 \\ 
			\hline 
		\end{tabular} 
	\end{center}
	\label{tab:ResSpecific}
\end{table}
\vspace{-5mm}

\subsection{Cyclists}
For the \textit{Cyclist Dataset}, the applicability of the PolyMLP concept to this VRU type could also be shown. As typical cyclist velocities at the test intersection are 2\,--\,3 times higher than those of pedestrians, their reach within the prediction horizon is much larger. Thus, the position prediction errors rise, here by 51\% for \textit{Starting} and for 79\% on \textit{Stopping} scenes compared to the \textit{Full Pedestrian Database}.
The fundamental insights to those gained for pedestrians: Especially regarding \textit{Starting} scenes, the PolyMLP model could outperform an optimized CV-KF by 46\%. Here, also physical modeling already leads to an improvement of 27\% compared to CV-KF. A significant advantage (34\%) of PolyMLP is also observed on stopping scenes, while the Kalman Filter prediction was not improved on \textit{Waiting} ($\pm$0\%) and \textit{Moving} scenes (-11\%). For a more detailed evaluation on the cyclist dataset, see \cite{Zernetsch16}. \vspace{-5mm}

\subsection{Processing Time}
Another important aspect is the computational time for training and for the online application of the predictor.
As the dimensionality of the in- and output features is relatively low due to the polynomial representation, the training time of the MLP also remains short. Using a current desktop PC (Intel Core i7-3770, 4\,$\times$\,3.4 GHz, 8\,GB RAM), 12\,min are needed for the single stage PolyMLP model, while 97\% of the final prediction quality is already reached after approx. 1\,min. 

For the two-stage approach, the preprocessing and the polynomial approximation step have only to be performed once as long as the same polynomial configuration is used for all models included. With regard to the one-stage approach, the computation time rises by only 35\% to 70\,$\mu$s. Altogether the algorithms perform very efficient with computational times several orders of magnitude lower than, e.\,g., most techniques of pedestrian or cyclist detection.
The used algorithms of video-based head detection and tracking, e.\,g., are processed within less than 40\,ms on the test system. Therefore, a HOG/SVM pedestrian detector processing full-frames on the GPU in order to deliver ROIs for the head detection is the limiting factor. \vspace{-3mm}

\section{Conclusion and Future Work} \label{conclusion}
	In this article we proposed VRU movement models based on machine learning methods. The presented approach uses the measured VRU trajectory to predict the current motion state and the future trajectory. It is compared to CV- and IMM-Kalman-Filtering, physically based models for starting and stopping and video-based motion classification with a MCHOG descriptor and SVMs.
	The results show that PolyMLP clearly outperforms Kalman Filtering for classification and trajectory prediction, in particular in the cases of starting and stopping motions. Physical models perform also better than KF, but are outperformed by PolyMLP too, especially considering the handling of false positive classifications. MCHOG/SVM uses additional image-based information as input and thus shows slightly more reliable early detection of starting motion (70\,ms faster on avrg.), but is approx. 400\,ms slower at the classification of stopping.
	The two-stage approach of motion state classification and trajectory prediction did not lead to a significant further improvement of the prediction quality in our case, but reveals the potential to include other classification of prediction models, perhaps basing on complementary sensors. 
	Altogether, PolyMLP represents a promising method for intention recognition of VRUs with state-of-the-art prediction quality and a high degree of flexibility with regard to VRU type, used sensor system, predicted time horizon and the concrete field of ADAS application.
	
	Our future work comprises the further evaluation of the techniques using on-board sensors in a moving vehicle under various traffic conditions. Here, especially the handling of measurement noise and the vehicle's ego motion represent additional challenges.
	Another promising approach is the inclusion of additional information, e.\,g. from high-precision maps, other on-board sensors, infrastructure knots or vehicles via V2I or V2V communication.  \vspace{-2mm}

\appendices


\section*{Acknowledgment}
This work partially results from the project AFUSS, supported by the German
Federal Ministry of Education and Research (BMBF) under grant number 03FH021I3, and the
project DeCoInt$^2$, supported by the German Research Foundation (DFG) within the priority program
SPP 1835: ``Kooperativ Interagierende Automobile'', grant numbers DO 1186/1-1 and SI 674/11-1. \vspace{-3mm}

\ifCLASSOPTIONcaptionsoff
  \newpage
\fi



\bibliographystyle{IEEEtran}
\bibliography{IEEEabrv,bibliography}
\vspace{-10mm}
%



%

\begin{IEEEbiography}[{\includegraphics[width=1in,height=1.25in,clip,keepaspectratio]{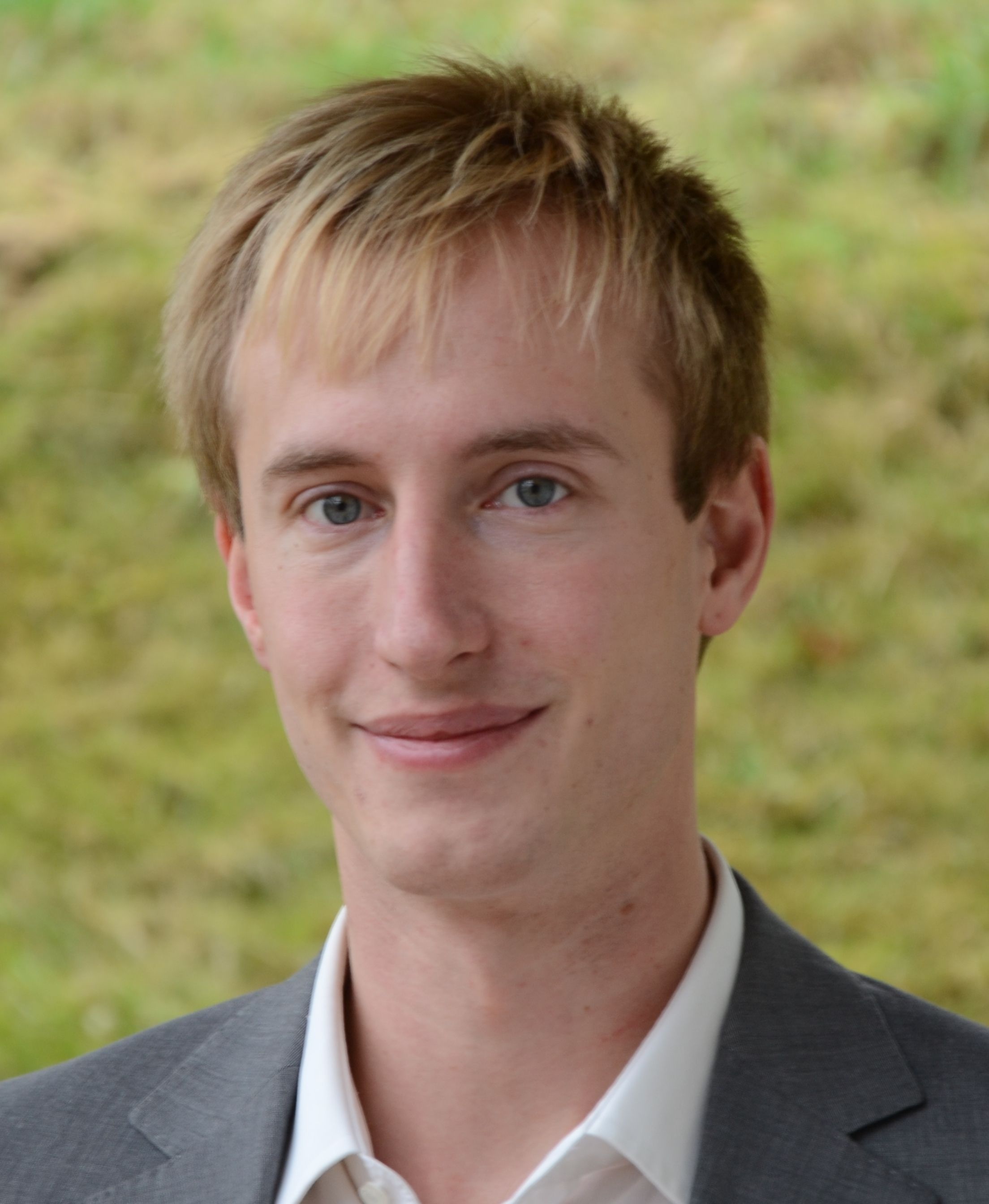}}]{Michael Goldhammer}
received the M. Eng. degree in Electrical Engineering and Information Technology from the University of Applied Sciences Aschaffenburg (Germany) in 2009 and the Dr. rer. nat. degree in computer sciences in 2016 from the University of Kassel (Germany). His PHD thesis focuses on self-learning algorithms for video-based intention detection of pedestrians. His research interests include machine vision and self-learning methods for sensor data processing and automotive safety purposes. 
\end{IEEEbiography}\vspace{-5mm}

\begin{IEEEbiography}[{\includegraphics[width=1in,height=1.25in,clip,keepaspectratio]{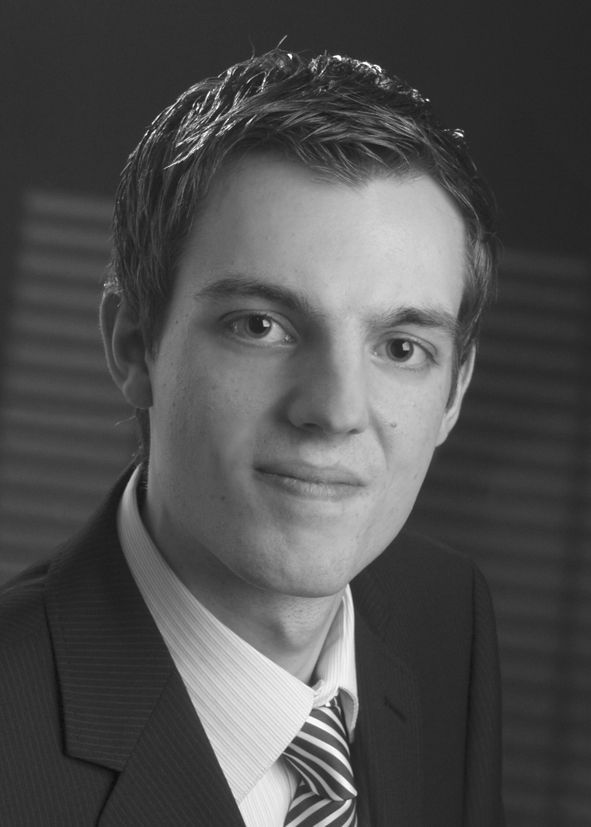}}]{Sebastian Köhler} received the Dipl.-Ing.~(FH) degree in Mechatronics and the M. Eng. degree in Electrical Engineering and Information Technology from the University of Applied Sciences Aschaffenburg, Germany, in 2010 and 2011, respectively. Currently, he is working on his PhD thesis in cooperation with the Institute of Measurement, Control, and Microtechnology, University of Ulm, Germany, focussing on intention detection of pedestrians in urban traffic. His main research interests include stereo vision, sensor and information fusion, road user detection and short-term behavior recognition for ADAS.
\end{IEEEbiography}


\begin{IEEEbiography}[{\includegraphics[width=1in,height=1.25in,clip,keepaspectratio]{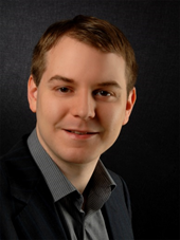}}]{Stefan Zernetsch}
received the B.Eng. and the M.Eng. degree in Electrical Engineering and Information Technology from the University of Applied Sciences Aschaffenburg, Germany, in 2012 and 2014, respectively. Currently, he is working on his PhD thesis in cooperation with the Faculty of Electrical Engineering and Computer Science of the University of Kassel, Germany. His research interests include cooperative sensor networks, sensor data fusion, multiple view geometry, pattern recognition and short-term behavior recognition of traffic participants.
\end{IEEEbiography}

\begin{IEEEbiography}[{\includegraphics[width=1in,height=1.25in,clip,keepaspectratio]{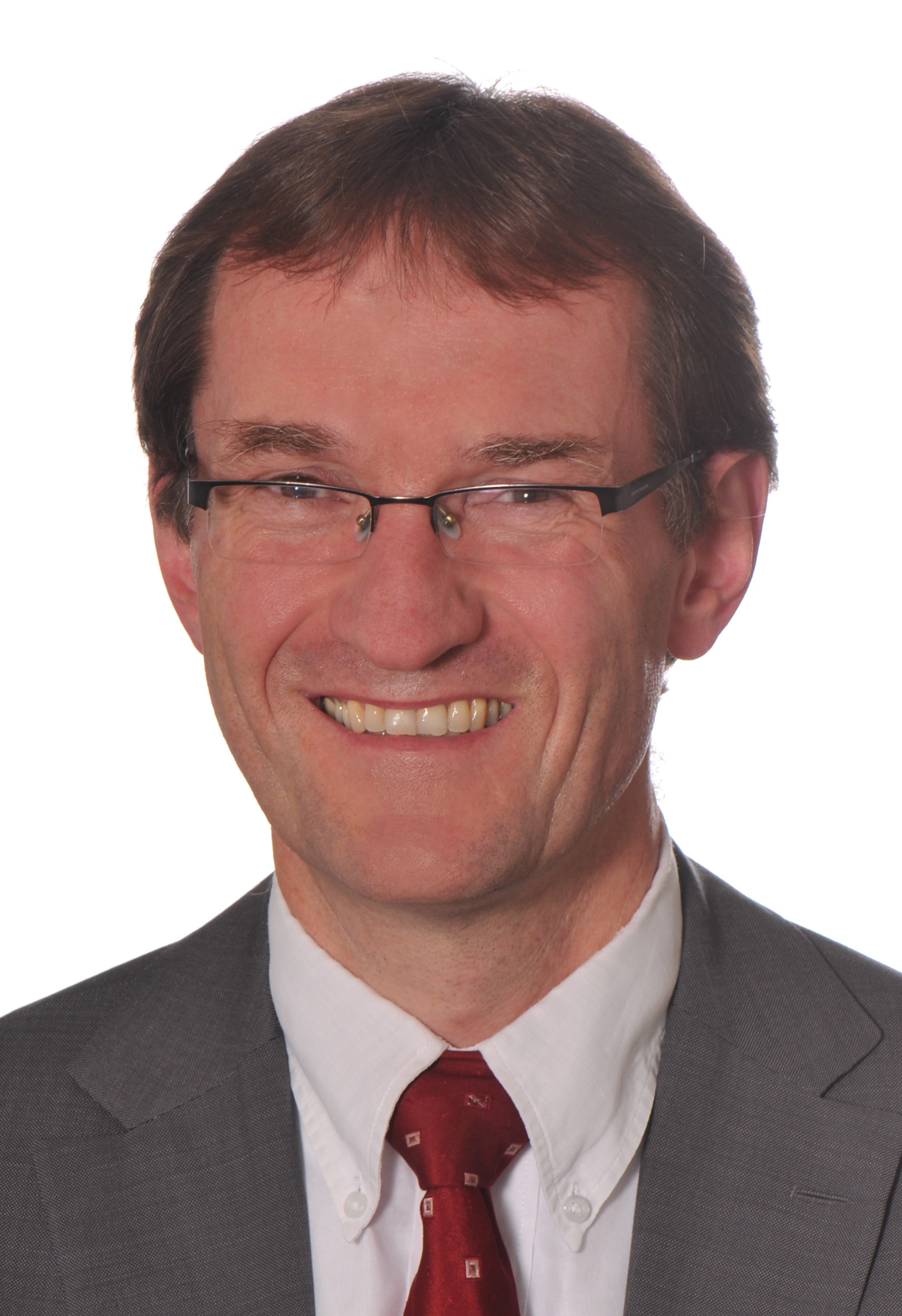}}]{Konrad Doll}
received the Diploma (Dipl.-Ing.) degree and the Dr.-Ing. degree in Electrical Engineering and Information Technology from the Technical University of Munich, Germany, in 1989 and 1994, respectively. In 1994 he joined the Semiconductor Products Sector of Motorola, Inc. (now Freescale Semiconductor, Inc.). In 1997 he was appointed to professor at the University of Applied Sciences Aschaffenburg in the field of computer science and digital systems design. His research interests include intelligent systems, their real-time implementations on platforms like CPU, GPUs and FPGAs and their applications in advanced driver assistance systems. Konrad Doll is member of the IEEE.
\end{IEEEbiography}

\begin{IEEEbiography}[{\includegraphics[width=1in,height=1.25in,clip,keepaspectratio]{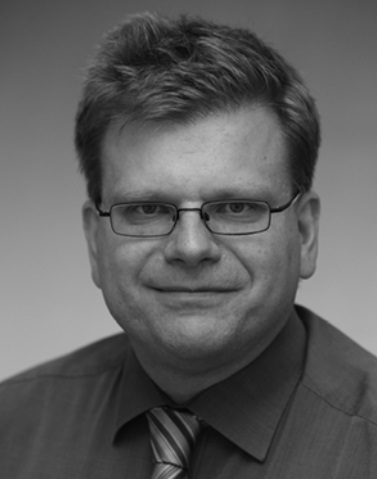}}]{Bernhard Sick}
received the diploma, the Ph.D. degree, and the "Habilitation" degree, all in computer science, from the University of Passau, Germany, in 1992, 1999, and 2004, respectively. Currently, he is full Professor for intelligent embedded systems at the Faculty for Electrical Engineering and Computer Science of the University of Kassel, Germany. There, he is conducting research in the areas autonomic and organic computing and technical data analytics with applications in biometrics, intrusion detection, energy management, automotive engineering, and others. He authored more than 90 peer-reviewed publications in these areas.\\
Dr. Sick is associate editor of the IEEE TRANSACTIONS ON CYBERNETICS. He holds one patent and received several thesis, best paper, teaching, and inventor awards. He is a member of IEEE (Systems, Man, and Cybernetics Society, Computer Society, and Computational Intelligence Society) and GI (Gesellschaft fuer Informatik).
\end{IEEEbiography}

\begin{IEEEbiography}[{\includegraphics[width=1in,height=1.25in,clip,keepaspectratio]{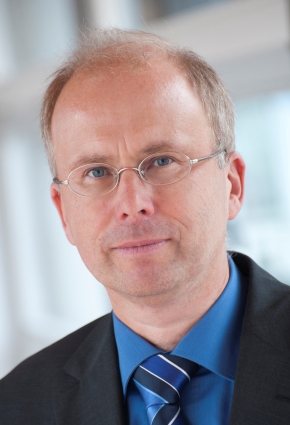}}]{Klaus Dietmayer}
was born in Celle, Germany in 1962. He received the Diploma degree (equivalent to M.Sc. degree) in electrical engineering
from Braunschweig University of Technology, Braunschweig, Germany, in 1989 and the Dr.-Ing. degree (equivalent to Ph.D. degree) from the Helmut
Schmidt University, Hamburg, Germany, in 1994. In 1994, he joined the Philips Semiconductors Systems
Laboratory, Hamburg, as a Research Engineer. Since 1996, he has been a Manager in the field of networks
and sensors for automotive applications. In 2000, he was appointed to a professorship at Ulm University, Ulm, Germany, in the field
of measurement and control. He is currently a Full Professor and the Director of the Institute of Measurement, Control and Microtechnology with the School
of Engineering and Computer Science, Ulm University. His research interests include information fusion, multiobject tracking, environment perception for advanced automotive driver assistance, and E-Mobility.\\
Dr. Dietmayer is a member of the German Society of Engineers VDI/VDE.
\end{IEEEbiography}


\vfill


\end{document}